\documentclass[11pt,a4paper]{article}
\usepackage[hyperref]{emnlp-ijcnlp-2019}
\usepackage{times}
\usepackage{latexsym}

\usepackage{url}

\usepackage{soul}
\setstcolor{red}
\usepackage[colorinlistoftodos,prependcaption,textsize=tiny]{todonotes}
\usepackage{url}

\usepackage{array,multirow,graphicx}
\usepackage{ifthen, color}

\newcommand{\wmtnmt}{$wmt18^{nmt}_{best}$}
\newcommand{\wmtnmtbold}{$\mathbf{wmt18^{nmt}_{best}}$}
\newcommand{\wmtsmt}{$wmt18^{smt}_{best}$}
\newcommand{\wmtsmtbold}{$\mathbf{wmt18^{smt}_{best}}$}
\newcommand{\wmtsmtgeneric}{$wmt18^{smt,generic}_{best}$}
\newcommand{\wmtsmtft}{$wmt18^{smt,ft}_{best}$}
\newcommand{\transfsmt}{$\{src,mt\}^{smt}_{tr} \rightarrow pe$}
\newcommand{\transfsmtbold}{$\mathbf{\{src,mt\}^{smt}_{tr} \rightarrow pe}$}
\newcommand{\transfnmt}[2]{$\{src,mt\}^{nmt}_{tr} \rightarrow pe^{{#1}}_{{#2}}$}
\newcommand{\transfnmtbold}[2]{$\mathbf{\{src,mt\}^{nmt}_{tr} \rightarrow pe^{{#1}}_{{#2}}}$}
\newcommand{\transf}{$\{src,mt\}_{tr} \rightarrow pe$}
\newcommand{\ensx}{$ensemble^{smt} (x3)$}
\newcommand{\ensxbold}{$\mathbf{ensemble^{smt} (x3)}$}
\newcommand{\ensckpt}{Exp3.3$^{smt}_{ens4ckpt}$}
\aclfinalcopy 

\title{The Transference Architecture for Automatic Post-Editing}

\author{Santanu Pal\textsuperscript{1,2}, Hongfei Xu\textsuperscript{1,2}, Nico Herbig\textsuperscript{2}, 
Sudip Kumar Naskar\textsuperscript{3},
Antonio Kr\"uger\textsuperscript{2}, \\
\textbf{Josef van Genabith\textsuperscript{1,2}}\\
  \textsuperscript{1}Department of Language Science and Technology,\\ Saarland University, Germany\\
  \textsuperscript{2}German Research Center for Artificial Intelligence (DFKI),\\ Saarland Informatics Campus, Germany
  \\\textsuperscript{3}Jadavpur University, Kolkata, India\\
  {\tt \{santanu.pal, josef.vangenabith\}@uni-saarland.de}\\ {\tt \{hongfei.xu, nico.herbig, krueger\}@dfki.de}, {\tt sudip.naskar@gmail.com} 
  \\
  }

\date{}

\begin{document}
\maketitle
\begin{abstract}
In automatic post-editing (APE) it makes sense to condition post-editing ($pe$) decisions on both the source ($src$) and the machine translated text ($mt$) as input. This has led to multi-source encoder based APE approaches. A research challenge now is the search for architectures that best support the capture, preparation and provision of $src$ and $mt$ information and its integration with $pe$ decisions. In this paper we present a new multi-source APE model, called \textit{transference}. Unlike previous approaches, it (i) uses a transformer encoder block for $src$, (ii) followed by a decoder block, but without masking for self-attention on $mt$, which effectively acts as second encoder combining  $src \rightarrow mt$, and (iii) feeds this representation into a final decoder block generating $pe$. Our model outperforms the state-of-the-art by 1 BLEU point on the WMT 2016, 2017, and 2018 English--German APE shared tasks (PBSMT and NMT). We further investigate the importance of our newly introduced second encoder and find that a too small amount of layers does hurt the performance, while reducing the number of layers of the decoder does not matter much.
\end{abstract}

\section{Introduction}
The performance of state-of-the-art MT systems is not perfect, thus, human interventions are still required to correct machine translated texts into publishable quality translations~\cite{tausb:2010}.
Automatic post-editing (APE) is a method that aims to automatically correct errors made by MT systems before performing actual human post-editing (PE)~\cite{Knight:1994:APE}, thereby reducing the translators' workload and increasing productivity~\cite{Pal:2016:COLING:MAIN}.
APE systems trained on human PE data serve as MT post-processing modules to improve the overall performance. 
APE can therefore be viewed as a 2\textsuperscript{nd}-stage MT system, translating predictable error patterns in MT output to their corresponding corrections. APE training data minimally involves MT output ($mt$) and the human post-edited ($pe$) version of $mt$, but additionally using the source ($src$) has been shown to provide further benefits~\cite{bojar-EtAl:2015:WMT,bojar-EtAl:2016:WMT, bojar-EtAl:2017:WMT}.

To provide awareness of errors in $mt$ originating from $src$, attention mechanisms~\cite{Bahdanau:2014:Neural} allow modeling of non-local dependencies in the input or output sequences, and importantly also global dependencies between them (in our case $src$, $mt$ and $pe$). 
The \textit{transformer} architecture~\cite{Vaswani:NIPS2017} is built solely upon such attention mechanisms completely replacing recurrence and convolutions.  
The transformer uses positional encoding to encode the input and output sequences, and computes both self- and cross-attention through so-called multi-head attentions, which are facilitated by parallelization.
Such multi-head attention allows to jointly attend to information at different positions from different representation subspaces, e.g.\ utilizing and combining information from $src$, $mt$, and $pe$.

In this paper, we present a multi-source neural APE architecture called \textit{transference}.
Our model contains a source encoder which encodes $src$ information, 
a second encoder ($enc_{src \rightarrow mt}$) which takes the encoded representation from the source encoder ($enc_{src}$), combines this with the self-attention-based encoding of $mt$ ($enc_{mt}$), and prepares a representation for the decoder ($dec_{pe}$) via cross-attention.
Our second encoder ($enc_{src \rightarrow mt}$) can also be viewed as a standard transformer decoding block, however, without masking, which acts as an encoder. We thus recombine the different blocks of the transformer architecture and repurpose them for the APE task in a simple yet effective way.
The suggested architecture is inspired by the two-step approach professional translators tend to use during post-editing: first, the source segment is compared to the corresponding translation suggestion (similar to what our $enc_{src \rightarrow mt}$ is doing), then corrections to the MT output are applied based on the encountered errors (in the same way that our $dec_{pe}$ uses the encoded representation of $enc_{src \rightarrow mt}$ to produce the final translation). 

The paper makes the following contributions: (i) we propose a new multi-encoder model for APE that consists only of standard transformer encoding and decoding blocks, (ii) by using a mix of self- and cross-attention we provide a representation of both $src$ and $mt$ for the decoder, allowing it to better capture errors in $mt$ originating from $src$; this advances the state-of-the-art in APE in terms of BLEU and TER, and (iii), we analyze the effect of varying the number of encoder and decoder layers \cite{domhan-2018-much}, indicating that the encoders contribute more than decoders in transformer-based neural APE.




\section{Related Research}
\label{sec:rl}

Recent advances in APE research are directed towards neural APE, which was first proposed by \newcite{Pal:2016:ACL} and \newcite{junczysdowmunt-grundkiewicz:2016:WMT} for the single-source APE scenario which does not consider $src$, i.e.\ $mt \rightarrow pe$. 
In their work, \newcite{junczysdowmunt-grundkiewicz:2016:WMT} also generated a large synthetic training dataset through back translation, which we also use as additional training data. 
%

Exploiting source information as an additional input can help neural APE to disambiguate corrections applied at each time step; this naturally leads to multi-source APE ($\{src, mt\} \rightarrow pe$).
A multi-source neural APE system can be configured either by using a single encoder that encodes the concatenation of $src$ and $mt$~\cite{niehues-EtAl:2016:COLING} or by using two separate encoders for $src$ and $mt$ and passing the concatenation of both encoders' final states to the decoder~\cite{Libovicky-EtAl:2016:WMT}. A few approaches to multi-source neural APE were proposed in the WMT 2017 APE shared task. \newcite{Junczysdowmunt:2017:WMT} 
combine both $mt$ and $src$ in a single neural architecture, exploring different combinations of attention mechanisms including soft attention and hard monotonic attention.
\newcite{Chatterjee-EtAl:2017:WMT2} built upon the two-encoder architecture of multi-source models~\cite{Libovicky-EtAl:2016:WMT} by means of concatenating both weighted contexts of encoded $src$ and $mt$. 
\newcite{Varis-bojar:2017:WMT} compared two multi-source models, one using a single encoder with concatenation of $src$ and $mt$ sentences, and a second one using two character-level encoders for $mt$ and $src$ along with a character-level decoder.

Recently, in the WMT 2018 APE shared task, several adaptations of the transformer architecture have been presented for multi-source APE
. \newcite{pal-EtAl:2018:WMT} proposed an APE model that uses three self-attention-based encoders. 
They introduce an additional joint encoder that attends over a combination of the two encoded sequences from $mt$ and $src$.  
\newcite{tebbifakhr-EtAl:2018:WMT}, the NMT-subtask winner of WMT 2018 (\wmtnmt{}), employ sequence-level loss functions in order to avoid exposure bias during training and to be consistent with the automatic evaluation metrics.
\newcite{shin-lee:2018:WMT} propose that each encoder 
has its own self-attention and feed-forward layer to process each input separately. On the decoder side, they add two additional multi-head attention layers, one for $src \rightarrow mt$ and another for $src \rightarrow pe$. Thereafter another multi-head attention between the output of those attention layers helps the decoder to capture common words in $mt$ which should remain in $pe$. The APE PBSMT-subtask winner of WMT 2018 (\wmtsmt{}) \cite{junczysdowmunt-grundkiewicz:2018:WMT} also presented another transformer-based multi-source APE which uses two encoders and stacks an additional cross-attention component for $src \rightarrow pe$ above the previous cross-attention for $mt \rightarrow pe$. Comparing \newcite{shin-lee:2018:WMT}'s approach with the winner system, there are only two differences in the architecture: (i) the cross-attention order of $src \rightarrow mt$ and $src \rightarrow pe$ in the decoder, and (ii) \wmtsmt{} additionally shares parameters between two encoders.

\section{Transference Model for APE}
\label{sec:mst-tr}
\begin{figure}[ht]
    \centering
    \includegraphics[scale=0.32]{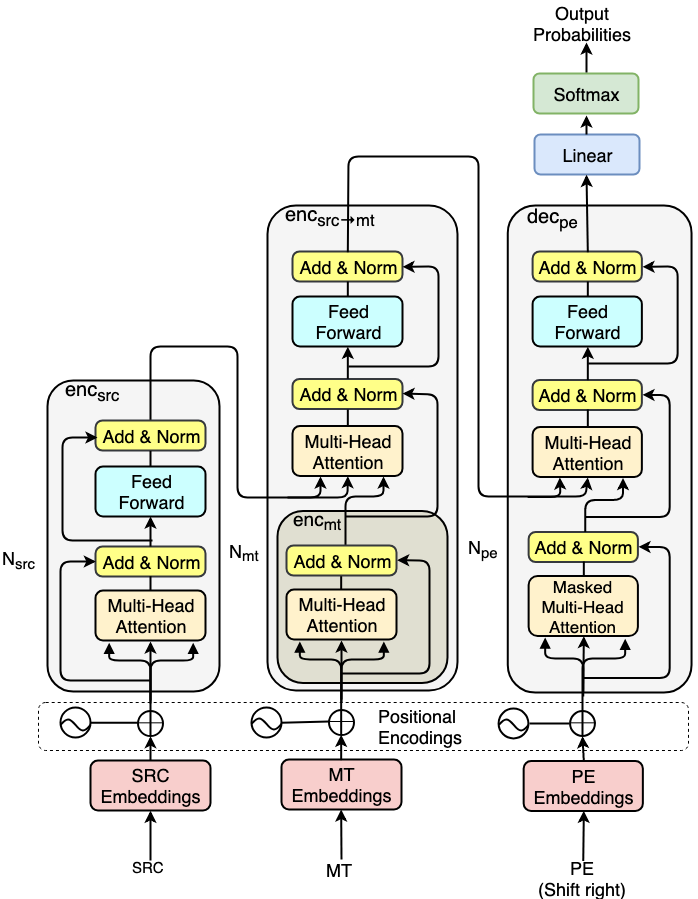}
    \caption{The \textit{transference} model architecture for APE (\transf{}).}
    \label{fig:mst_inferred}
\end{figure}

We propose a multi-source transformer model called \textit{transference} (\transf{}, Figure~\ref{fig:mst_inferred}), which takes advantage of both the encodings of $src$ and $mt$ and attends over a combination of both sequences while generating the post-edited sentence. The second encoder, $enc_{src \rightarrow mt}$, makes use of the first encoder $enc_{src}$ and a sub-encoder $enc_{mt}$ for considering $src$ and $mt$. Here, the $enc_{src}$ encoder and the $dec_{pe}$ decoder are equivalent to the original transformer for neural MT.  
Our $enc_{src \rightarrow mt}$ follows an architecture similar to the transformer's decoder, the difference being that no masked multi-head self-attention is used to process $mt$. 

One self-attended encoder for $src$, $\mathbf{s}$ = $(s_1, s_2, \ldots, s_k)$, returns a sequence of continuous representations, $enc_{src}$, and a second self-attended sub-encoder for $mt$, $\mathbf{m}$ = $(m_1, m_2, \ldots, m_l)$, returns another sequence
of continuous representations, $enc_{mt}$. Self-attention at this point provides the advantage of aggregating information from all of the words, including $src$ and $mt$, and successively generates a new representation per word informed by the entire $src$ and $mt$ context. The internal $enc_{mt}$ representation performs cross-attention over $enc_{src}$ and prepares a final representation ($enc_{src \rightarrow mt}$) for the decoder ($dec_{pe}$).  
The decoder then generates the $pe$ output in sequence, $\mathbf{p}$ = $(p_1, p_2, \ldots, p_n)$, one word at a time from left to right by attending to previously generated words as well as the final representations ($enc_{src \rightarrow mt}$) generated by the encoder.

To summarize, our multi-source APE implementation extends \newcite{Vaswani:NIPS2017} by introducing an additional encoding block by which $src$ and $mt$ communicate with the decoder. 

Our proposed approach differs from the WMT 2018 PBSMT winner system in several ways: (i) we use the original transformer's decoder without modifications; (ii) one of our encoder blocks ($enc_{src \rightarrow mt}$) is identical to the transformer's decoder block but uses no masking in the self-attention layer, thus having one self-attention layer and an additional cross-attention for $src \rightarrow mt$; and (iii) in the decoder layer, the cross-attention is performed between the encoded representation from $enc_{src \rightarrow mt}$ and $pe$. 

Our approach also differs from the WMT 2018 NMT winner system: (i) \wmtnmt{} concatenates the encoded representation of two encoders and passes it as
the key to the attention layer of the decoder, and (ii), the system additionally employs sequence-level loss functions based on maximum likelihood estimation and minimum risk training in order to
avoid exposure bias during training. 


The main intuition is that our $enc_{src  \rightarrow mt}$ attends over the $src$ and $mt$ and informs the $pe$ to better capture, process, and share information between $src$-$mt$-$pe$, which efficiently models error patterns and the corresponding corrections. Our model performs better than past approaches, as the experiment section will show.

\section{Experiments}
We explore our approach on both APE sub-tasks of WMT 2018, where the 1\textsuperscript{st}-stage MT system to which APE is applied is either a phrase-based statistical machine translation (PBSMT) or a neural machine translation (NMT) model.

For the PBSMT task, we compare against four baselines: the \textbf{raw SMT} output provided by the 1\textsuperscript{st}-stage PBSMT system, the best-performing systems from WMT APE 2018 (\wmtsmtbold{}), which are a single model and an ensemble model by~\newcite{junczysdowmunt-grundkiewicz:2018:WMT}, as well as a transformer trying to directly translate from $src$ to $pe$ (\textbf{Transformer} ($\mathbf{src \rightarrow pe}$)), thus performing translation instead of APE. We evaluate the systems using BLEU \cite{Papineni:2002} and TER \cite{Snover:2006}.

For the NMT task, we consider two baselines: the \textbf{raw NMT} output provided by the 1\textsuperscript{st}-stage NMT system and the best-performing system from the WMT 2018 NMT APE task (\wmtnmtbold{}) \cite{tebbifakhr-EtAl:2018:WMT}.

Apart from the multi-encoder \textit{transference} architecture described above (\transf{}) and ensembling of this architecture, two simpler versions are also analyzed: first, a `mono-lingual' ($\mathbf{mt \rightarrow pe}$) APE model using only parallel $mt$--$pe$ data and therefore only a single encoder, and second, an identical single-encoder architecture, however, 
using the concatenated $src$ and $mt$ text as input 
($\mathbf{\{src+mt\} \rightarrow pe}$)~\cite{niehues-EtAl:2016:COLING}. 

\subsection{Data}
\label{dataset}
For our experiments, we use the English--German WMT 2016~\cite{bojar-EtAl:2016:WMT}, 2017~\cite{bojar-EtAl:2017:WMT} and 2018~\cite{chatterjee-EtAl:2018:WMTAPE} APE task data. 
All these released APE datasets consist of English--German triplets containing source English text ($src$) from the IT domain, the corresponding German translations ($mt$) from a 1\textsuperscript{st}-stage MT system, and the corresponding human-post-edited version ($pe$). 
The sizes of the datasets (train; dev; test), in terms of number of sentences, are (12,000; 1,000; 2,000), (11,000; 0; 2,000), and (13,442; 1,000; 1,023), for the 2016 PBSMT, the 2017 PBSMT, and the 2018 NMT data, respectively. One should note that for WMT 2018, we carried out experiments only for the NMT sub-task and ignored the data for the PBSMT task.

Since the WMT APE datasets are small in size, we use `artificial training data'~\cite{junczysdowmunt-grundkiewicz:2016:WMT} containing 4.5M sentences as additional resources, 4M of which are weakly similar to the WMT 2016 training data, while 500K are very similar according to TER statistics. 

For experimenting on the NMT data, we additionally use the synthetic \mbox{eScape} APE corpus \cite{eSCAPE:2018}, consisting of $\sim$7M triples.
For cleaning this noisy \mbox{eScape} dataset containing many unrelated language words (e.g.\ Chinese), we perform the following two steps: (i) we use the cleaning process described in \newcite{tebbifakhr-EtAl:2018:WMT}, and (ii) we use the Moses~\cite{Koehn:2007} corpus cleaning scripts with minimum and maximum number of tokens set to 1 and 100, respectively. After cleaning, we perform punctuation normalization, and then use the Moses tokenizer~\cite{Koehn:2007} to tokenize the \mbox{eScape} corpus with `no-escape' option. Finally, we apply true-casing. The cleaned version of the \mbox{eScape} corpus contains $\sim$6.5M triplets.

	

\subsection{Experiment Setup}
\label{sec:exp_setup}
To build models for the PBSMT tasks from 2016 and 2017, we first train a generic APE model using all the training data (4M + 500K + 12K + 11K) described in Section \ref{dataset}.
Afterwards, we fine-tune the trained model using the 500K artificial and 23K (12K + 11K) real PE training data.  
We use the WMT 2016 development data (dev2016) containing 1,000 triplets to validate the models during training.
To test our system performance, we use the WMT 2016 and 2017 test data (test2016, test2017) as two sub-experiments, each containing 2,000 triplets ($src$, $mt$ and $pe$).  We compare the performance of our system with the four different baseline systems described above: raw MT, \wmtsmt{} single and ensemble, as well as Transformer ($src \rightarrow pe$).

Additionally, we check the performance of our model on the WMT 2018 NMT APE task (where unlike in previous tasks, the 1\textsuperscript{st}-stage MT system is provided by NMT): for this, we explore two experimental setups: (i) we use the PBSMT task's APE model as a generic model which is then fine-tuned to a subset (12k) of the NMT data (\transfnmt{generic, smt}{}). One should note that it has been argued that the inclusion of SMT-specific data could be harmful when training NMT APE models~\cite{junczysdowmunt-grundkiewicz:2018:WMT}. (ii), we train a completely new generic model on the cleaned eScape data ($\sim$6.5M) along with a subset (12K) of the original training data released for the NMT task (\transfnmt{generic, nmt}{}). The aforementioned 12K NMT data are the first 12K of the overall 13.4K NMT data. The remaining 1.4K are used as validation data. The released development set (dev2018) is used as test data for our experiment, alongside the test2018, for which we could only obtain results for a few models by the WMT 2019 task organizers. 
We also explore an additional fine-tuning step of \transfnmt{generic, nmt}{} towards the 12K NMT data (called \transfnmt{ft}{}), and a model averaging the 8 best checkpoints of \transfnmt{ft}{}, which we call \transfnmt{ft}{avg}.

Last, we analyze the importance of our 
second encoder ($enc_{src \rightarrow mt}$), compared to the source encoder ($enc_{src}$) and the decoder ($dec_{pe}$), by reducing and expanding the amount of layers in the encoders and the decoder. Our standard setup, which we use for fine-tuning, ensembling etc., is fixed to 6-6-6 for $N_{src}$-$N_{mt}$-$N_{pe}$ (cf.\ Figure~\ref{fig:mst_inferred}), where 6 is the value that was proposed by \newcite{Vaswani:NIPS2017} for the \textit{base} model.
We investigate what happens in terms of APE performance if we change this setting to 6-6-4 and 6-4-6.

To handle out-of-vocabulary words and reduce the vocabulary
size, instead of considering words, we consider subword units~\cite{Sennrich:2016ACL} by using byte-pair encoding (BPE). In the preprocessing step, instead of learning an explicit mapping between BPEs in the
$src$, $mt$ and $pe$, we define BPE tokens by jointly processing all triplets. Thus, $src$, $mt$ and $pe$ derive a single BPE vocabulary.
Since $mt$ and $pe$ belong to the same language (German) and $src$ is a close language (English), they naturally share a good fraction of BPE tokens, which reduces the vocabulary size to 28k. 


\subsection{Hyper-parameter Setup}
We follow a similar hyper-parameter setup for all reported systems.
All encoders (for $\{src,mt\}_{tr} \rightarrow pe$), and the decoder, are composed of a stack of $N_{src} = N_{mt} = N_{pe} = 6$ identical layers followed by layer normalization. 
The learning rate is varied throughout the training process, and increasing for the first training steps $warmup_{steps} = 8000$ and afterwards decreasing as described in~\cite{Vaswani:NIPS2017}. All remaining hyper-parameters are set analogously to those of the transformer's \textit{base} model, except that we do not perform checkpoint averaging.
%
At training time, the batch size is set to 25K tokens, with a maximum sentence length of 256 subwords. 
After each epoch, the training data is shuffled. 
During decoding, we perform beam search with a beam size of 4. We use shared embeddings between $mt$ and $pe$ in all our experiments.

\begin{table*}[htbp]
\small
\centering
\begin{tabular}{l|l||ll||ll}
\multirow{2}{*}{\begin{tabular}[c]{@{}l@{}}Exp.\\ no.\end{tabular}}   & \multirow{2}{*}{Models} & \multicolumn{2}{c||}{test2016} & \multicolumn{2}{c}{test2017}  \\
\cline{3-6}
                     &   & BLEU $\uparrow$  & TER $\downarrow$           & BLEU $\uparrow$  & TER $\downarrow$                    \\
                     \hline
\multicolumn{6}{l}{\textbf{Baselines}} \\
\hline
1.1 & Raw SMT &    62.11               &     24.76          &    62.49             &   24.48                \\
1.2 & Transformer ($src \rightarrow pe$)  &    56.59 (-5.52)              &     29.97 (+5.21)         &    53.06 (-9.43)             &   32.20 (+7.72)              \\
1.3 & \wmtsmt{} (single) & 70.86 (+8.75) & 18.92 (-5.84) & 69.72 (+7.23) & 19.49 (-4.99) \\
1.4  & \wmtsmt{} (x4)  & \textbf{71.04} (+8.93)     &     \textbf{18.86} (-5.9)          &     \textbf{70.46} (+7.97)            &     \textbf{19.03} (-5.45)          \\
\cline{2-6}
& \multicolumn{5}{l}{\textbf{Baselines: Retrained \wmtsmt{} with our experimental setup}} \\
\cline{2-6}
1.5 & \wmtsmtgeneric{} (single) & 69.14 (+7.03) & 20.41 (-4.35)  & 68.14 (+5.65) &  20.98 (-3.5) \\
1.6 & \wmtsmtft{} (single) & 70.12 (+8.01) & 19.84 (-4.92) & 69.16 (+6.67)  & 20.34 (-4.14)  \\
                     \hline
                     
 \multicolumn{6}{l}{\textbf{General models trained on 23K+4.5M data}}                                              \\
\hline
                     
2.1 & $mt \rightarrow pe$    &    67.70 (+5.59)            &     21.90 (-2.86)        &          66.91 (+4.42)      &      22.32 (-2.16)                \\
2.2 & $\{src+mt\} \rightarrow pe$    & 69.32 (+7.21)      &      20.27 (-4.49)  &  68.26 (+5.77)               &     20.90 (-3.58)         \\
2.3 & \transfsmt{}  & 70.46 (+8.35)             &   19.21 (-5.55)            &   70.05 (+7.56)              &     19.46 (-5.02)                 \\
                     
                     \hline
\multicolumn{6}{l}{\textbf{Fine-tuning Exp.\ 2 models with 23K+500K data }}                                              \\
                     \hline
 3.1 & $mt \rightarrow pe$    &    68.43 (+6.32)              &    21.29 (-3.47)        &         67.78 (+5.29)        &       21.63 (-2.85)            \\
3.2 & $\{src+mt\} \rightarrow pe$  &  69.87 (+7.76)               &   19.94 (-4.82)            & 68.57 (+6.08)                &     20.68 (-3.8)              \\
3.3 & \transfsmt{}  &    71.05  (+8.94)   &   19.05 (-5.71)  &   70.33 (+7.84)  &    19.23 (-5.25)                \\
                     \hline

4.1 & \ensckpt   &     \textbf{71.59} (+9.48)  &     \textbf{18.78} (-5.98)  &     \textbf{70.89} (+8.4)    &     \textbf{18.91}  (-5.57)        \\
4.2 & \ensx{}     &   \textbf{72.19} (+10.08)     &  \textbf{18.39}  (-6.37)    &  \textbf{71.58}  (+9.09)          &     \textbf{18.58} (-5.9)         \\
                     \hline
\multicolumn{6}{l}{\textbf{\transfsmt{} with different layer size }} \\
\hline
5.1 & \transfsmt{} (6-6-4)                                                                                  &  70.85 (+8.74)	 &	 19.00 (-5.76)	& 69.82 (+7.33)	&         19.67 (-4.81)       \\
5.2 & \transfsmt{} (6-4-6)                                                                                   &    69.93 (+7.82)               &     19.70 (-5.06)         &    69.61 (+7.12)             &       19.68 (-4.8)           \\
\hline                     
                     
\end{tabular}
\caption{Evaluation results on the WMT APE test set 2016, and test set 2017 for the PBSMT task; ($\pm X$) value is the improvement over \wmtsmt{} (x4). The last section of the table shows the impact of increasing and decreasing the depth of the encoders and the decoder.}
\label{tab:smt-result}
\end{table*}

\section{Results}
\label{result}
The results of our four models, \textbf{\emph{single-source}} ($\mathbf{mt \rightarrow pe}$), \textbf{\emph{multi-source single encoder}} ($\mathbf{\{src + pe\} \rightarrow pe}$), \textbf{\emph{transference}} (\transfsmtbold{}), and \textbf{\emph{ensemble}}, in comparison to the four baselines, \textbf{\emph{raw SMT}}, \wmtsmtbold{}~\cite{junczysdowmunt-grundkiewicz:2018:WMT} single and ensemble, as well as \textbf{\emph{Transformer}} ($\mathbf{src \rightarrow pe}$), are presented in Table~\ref{tab:smt-result} for test2016 and test2017. Table~\ref{tab:nmt-result} reports the results obtained by our \textbf{\emph{transference}} model (\transfnmtbold{}{}) on the WMT 2018 NMT data for dev2018 (which we use as a test set) and test2018, compared to the baselines \textbf{\emph{raw NMT}} and \wmtnmtbold{}.

\subsection{Baselines}
The \textbf{raw SMT} output in Table~\ref{tab:smt-result} is a strong black-box PBSMT system (i.e., 1st-stage MT). We report its performance observed with respect to the ground truth ($pe$), i.e., the post-edited version of $mt$. The original PBSMT system scores over 62 BLEU points and below 25 TER on test2016 and test2017.

Using a \textbf{Transformer ($src \rightarrow pe$)}, we test if APE is really useful, or if potential gains are only achieved due to the good performance of the transformer architecture. While we cannot do a full training of the transformer on the data that the raw MT engine was trained on due to the unavailability of the data, we use our PE datasets in an equivalent experimental setup as for all other models. The results of this system (Exp.\ 1.2 in Table~\ref{tab:smt-result}) show that the performance is actually lower across both test sets, -5.52/-9.43 absolute points in BLEU and +5.21/+7.72 absolute in TER, compared to the raw SMT baseline.

We report four results from \wmtsmtbold{}, (i) \wmtsmt{} ($single$), which is the core multi-encoder implementation without ensembling but with checkpoint averaging, (ii) \wmtsmt{} ($x4$) which is an ensemble of four identical `single' models trained with different random initializations. The results of  \wmtsmt{} ($single$) and \wmtsmt{} ($x4$) (Exp.\ 1.3 and 1.4) reported in Table \ref{tab:smt-result} are from \newcite{junczysdowmunt-grundkiewicz:2018:WMT}.
Since their training procedure slightly differs from ours, we also trained the \wmtsmt{} system using exactly our experimental setup in order to make a fair comparison. This yields the baselines (iii) \wmtsmtgeneric{} ($single$) (Exp.\ 1.5), which is similar to \wmtsmt{} ($single$), however, the training parameters and data are kept in line with our \textit{transference} general model (Exp.\ 2.3) and (iv) \wmtsmtft{} ($single$) (Exp.\ 1.6), which is also trained maintaining the equivalent experimental setup compared to the fine tuned version of the \textit{transference} general model (Exp.\ 3.3). 
Compared to both raw SMT and Transformer ($src \rightarrow pe$) we see strong improvements for this state-of-the-art model, with BLEU scores of at least 68.14 and TER scores of at most 20.98 across the PBSMT testsets. \wmtsmt{}, however, performs better in its original setup (Exp.\ 1.3 and 1.4) compared to our experimental setup (Exp.\ 1.5 and 1.6).

\subsection{Single-Encoder Transformer for APE}
The two transformer architectures $\mathbf{mt \rightarrow pe}$ and $\mathbf{\{src+mt\} \rightarrow pe}$ use only a single encoder. Table \ref{tab:smt-result} shows that $\mathbf{mt \rightarrow pe}$ (Exp.\ 2.1) provides better performance (+4.42 absolute BLEU on test2017) compared to the original SMT, while $\mathbf{\{src+mt\} \rightarrow pe}$ (Exp.\ 2.2) provides further improvements by additionally using the $src$ information. $\mathbf{\{src+mt\} \rightarrow pe}$ improves over $\mathbf{mt \rightarrow pe}$ by +1.62/+1.35 absolute BLEU points on test2016/test2017. After fine-tuning, both single encoder transformers (Exp.\ 3.1 and 3.2 in Table \ref{tab:smt-result}) show further improvements, +0.87 and +0.31 absolute BLEU points, respectively, for test2017 and a similar improvement for test2016.

\subsection{Transference Transformer for APE}
In contrast to the two models above, our \textit{transference} architecture uses multiple encoders.
To fairly compare to \wmtsmt{}, we retrain the \wmtsmt{} system with our experimental setup (cf. Exp.\ 1.5 and 1.6 in Table \ref{tab:smt-result}). \wmtsmtgeneric{} (\textit{single}) is a generic model trained on all the training data; which is afterwards fine-tuned with 500K artificial and 23K real PE data (\wmtsmtft{} (\textit{single})). It is to be noted that in terms of performance the data processing method described in \newcite{junczysdowmunt-grundkiewicz:2018:WMT} reported in Exp.\ 1.3 is better than ours (Exp.\ 1.6). 
The fine-tuned version of the \transfsmt{} model (Exp.\ 3.3 in Table~\ref{tab:smt-result}) outperforms \wmtsmt{} (single) (Exp.\ 1.3) in BLEU on both test sets, however, the TER score for test2016 increases. 
One should note that \wmtsmt{} (single) follows the transformer \textit{base} model, which is an average of five checkpoints, while our Exp.\ 3.3 is not.
When ensembling the 4 best checkpoints of our \transfsmt{} model (Exp.\ 4.1), the result beats the \wmtsmt{} (x4) system, which is an ensemble of four different randomly initialized \wmtsmt{} (single) systems. Our \ensxbold{} combines two \transfsmt{} (Exp.\ 2.3) models initialized with different random weights with the ensemble of the fine-tuned transference model \ensckpt (Exp.\ 4.1). This ensemble provides the best results for all datasets, providing roughly +1 BLEU point and -0.5 TER when comparing against \wmtsmt{} (x4).

\begin{table*}[ht]
\centering
\small
\begin{tabular}{l|l||ll|ll}
 \multirow{2}{*}{\begin{tabular}[c]{@{}l@{}}Exp.\\ no.\end{tabular}} & \multirow{2}{*}{Models} &\multicolumn{2}{c}{dev2018}  &\multicolumn{2}{c}{test2018}\\
 \cline{3-6}

                 &  & BLEU $\uparrow$        & TER  $\downarrow$        & BLEU $\uparrow$        & TER  $\downarrow$            \\
                     \hline
                6.1   &  Raw NMT  & 76.76 & 15.08 & 74.73 & 16.80      \\
                6.2   &  \wmtnmt{}  & \textbf{77.74 (+0.98)} & 14.78 (-0.30)  & 75.53 (+0.80) &  16.46 (-0.30)   \\
                      
                      \hline
                \multicolumn{4}{l}{\textbf{Fine-tuning Exp.\ 3.3 on 12k NMT data}}  \\
                  \hline
                7    & \transfnmt{generic, smt}{}  &    77.09 (+0.33)  &  14.94 (-0.14) & - & -   \\
                \hline
                \multicolumn{4}{l}{\textbf{Transference model trained on eScape+ 12k NMT data}}  \\
                  \hline
                8 & \transfnmt{generic, nmt}{}&    77.25 (+0.49)   &  14.87 (-0.21)   & - & - \\
                \hline
                \multicolumn{4}{l}{\textbf{Fine-tuning model 8 on 12k NMT data}}  \\
                  \hline
                9    & \transfnmt{ft}{}\   &    77.39 (+0.63)    &   14.71 (-0.37)  & - & - \\
                     \hline 
                  \multicolumn{4}{l}{\textbf{Averaging 8 checkpoints of Exp.\ 9}}  \\
                  \hline
                10    & \transfnmt{ft}{avg} &    77.67 (+0.91)    &  \textbf{14.52 (-0.56)}  &    \textbf{75.75 (+1.02)}   &  \textbf{16.15  (-0.69)}    \\
                     
                     \hline
\end{tabular}
\caption{Evaluation results on the WMT APE 2018 development set for the NMT task (Exp.\ 10 results were obtained by the WMT 2019 task organizers).
}
\label{tab:nmt-result}
\end{table*}

The results on the WMT 2018 NMT datasets (dev2018 and test2018) are presented in Table~\ref{tab:nmt-result}.
The \textit{raw NMT} system serves as one baseline against which we compare the performance of the different models.
We evaluate the system hypotheses with respect to the ground truth ($pe$), i.e., the post-edited version of $mt$. The baseline original NMT system scores 76.76 BLEU points and 15.08 TER on dev2018, and 74.73 BLEU points and 16.84 TER on test2018.


For the WMT 2018 NMT data we first test our \transfnmt{generic,smt}{} model, which is the model from Exp.\ 3.3 fine-tuned towards NMT data as described in Section~\ref{sec:exp_setup}. Table~\ref{tab:nmt-result} shows that our PBSMT APE model fine-tuned towards NMT (Exp.\ 7) can even slightly improve over the already very strong NMT system by about +0.3 BLEU and -0.1 TER, although these improvements are not statistically significant.

The overall results improve when we train our model on eScape and NMT data instead of using the PBSMT model as a basis. Our proposed generic \textit{transference} model (Exp.\ 8, \transfnmt{generic,nmt}{} shows statistically significant improvements in terms of BLEU and TER compared to the baseline even before fine-tuning, and further improvements after fine-tuning (Exp.\ 9, \transfnmt{ft}{}). 
Finally, after averaging the 8 best checkpoints, our \transfnmt{ft}{avg} model (Exp.\ 10) also shows consistent improvements in comparison to the baseline and other experimental setups.
Overall our fine-tuned model averaging the 8 best checkpoints achieves +1.02 absolute BLEU points and -0.69 absolute TER improvements over the baseline on test2018.
Table~\ref{tab:nmt-result} also shows the performance of our  model compared to the winner system of WMT 2018 (\wmtnmt{}) for the NMT task \cite{tebbifakhr-EtAl:2018:WMT}.
\wmtnmt{} scores 14.78 in TER and 77.74 in BLEU on the dev2018 and 16.46 in TER and 75.53 in BLEU on the  test2018. In comparison to \wmtnmt{}, our model (Exp.\ 10) achieves better scores in TER on both the dev2018 and test2018, however, in terms of BLEU our model scores slightly lower for dev2018, while some improvements are achieved on test2018.

The number of layers ($N_{src}$-$N_{mt}$-$N_{pe}$) in all encoders and the decoder for these results is fixed to 6-6-6. In Exp.\ 5.1, and 5.2 in Table~\ref{tab:smt-result}, we see the results of changing this setting to 6-6-4 and 6-4-6. 
This can be compared to the results of Exp.\ 2.3, since no fine-tuning or ensembling was performed for these three experiments. 
Exp.\ 5.1 shows that decreasing the number of layers on the decoder side does not hurt the performance. In fact, in the case of test2016, we got some improvement, while for test2017, the scores got slightly worse. 
In contrast, reducing the $enc_{src \rightarrow mt}$ encoder block's depth (Exp.\ 5.2) does indeed reduce the performance for all four scores, showing the importance of this second encoder.

\subsection{Analysis of Error Patterns}


In Table~\ref{tab:ed_base}, we analyze and compare the best performing SMT (\ensx{}) and NMT (\transfnmt{ft}{avg}) model outputs with the original MT outputs on the WMT 2017 (SMT) APE test set and on the WMT 2018 (NMT) development set. Improvements are measured in terms of number of words which need to be (i) inserted (\textit{In}), (ii) deleted (\textit{De}), (iii) substituted (\textit{Su}), and (iv) shifted (\textit{Sh}), as per TER \cite{Snover:2006}, in order to turn the MT outputs into reference translations. 
Our model provides promising results by significantly reducing the required number of edits (24\% overall for PBSMT task and 3.6\% for NMT task) across all edit operations, thereby leading to reduced post-editing effort and hence improving human post-editing productivity. 

\begin{table}
\small
    \centering
    \begin{tabular}{l||rrrr}
            & \%\textit{In} & \%\textit{De} & \%\textit{Su} & \%\textit{Sh}\\
         \hline
       \begin{tabular}[c]{@{}l@{}}\ensx{} \\ vs.\ \textit{raw SMT} \end{tabular} & +31 & +29 & +15  & +32 \\
       \hline
       \begin{tabular}[c]{@{}l@{}} \transfnmt{ft}{avg} \\ vs.\ \textit{raw NMT} \end{tabular} & +6 & +2 & +4 & -2\\
       \hline
       
    \end{tabular}
    \caption{\% of error reduction in terms of different edit operations achieved by our best systems compared to the raw MT baselines.}
    \label{tab:ed_base}
\end{table}

When comparing PBSMT to NMT, we see that stronger improvements are achieved for PBSMT, probably because the raw SMT is worse than the raw NMT. For PBSMT, similar results are achieved for \textit{In}, \textit{De}, and \textit{Sh}, while less gains are obtained in terms of \textit{Su}. For NMT, \textit{In} is improved most, followed by \textit{Su}, \textit{De}, and last \textit{Sh}. For shifts in NMT, the APE system even creates further errors, instead of reducing them, which is an issue we aim to prevent in the future.


\subsection{Discussion}

The proposed \textit{transference} architecture (\transfsmt{}, Exp.\ 2.3) shows slightly worse results than \wmtsmt{} (single) (Exp.\ 1.3) before fine-tuning, and roughly similar results after fine-tuning (Exp.\ 3.3). 
After ensembling, however, our \textit{transference} model (Exp.\ 4.2) shows consistent improvements when comparing against the best baseline ensemble \wmtsmt{} (x4) (Exp.\ 1.4).
Due to the unavailability of the sentence-level scores of \wmtsmt{} (x4), we could not test if the improvements (roughly +1 BLEU, -0.5 TER) are statistically significant.
Interestingly, our approach of taking the model optimized for PBSMT and fine-tuning it to the NMT task (Exp.\ 7) does not hurt the performance as was reported in the previous literature~\cite{junczysdowmunt-grundkiewicz:2018:WMT}. In contrast, some small, albeit statistically insignificant improvements over the raw NMT baseline were achieved. 
When we train the \textit{transference} architecture directly for the NMT task (Exp.\ 8), we get slightly better and statistically significant improvements compared to raw NMT. Fine-tuning this NMT model further towards the actual NMT data (Exp.\ 9), as well as performing checkpoint averaging using the 8 best checkpoints improves the results even further.

The reasons for the effectiveness of our approach can be summarized as follows. 
(1) Our $enc_{src \rightarrow mt}$ contains two attention mechanisms: one is self-attention and another is cross-attention. The self-attention layer is not masked here; therefore, the cross-attention layer in $enc_{src \rightarrow mt}$ is informed by both previous and future time-steps from the self-attended representation of $mt$ ($enc_{mt}$) and additionally from $enc_{src}$. As a result, each state representation of $enc_{src \rightarrow mt}$ is learned from the context of $src$ and $mt$. This might produce better representations for $dec_{pe}$ which can access the combined context. In contrast, in \wmtsmt{}, the $dec_{pe}$ accesses representations from $src$ and $mt$ independently, first using the representation from $mt$ and then using that of $src$. 
(2) The position-wise feed-forward layer in our $enc_{src \rightarrow mt}$ of the \textit{transference} model requires processing information from two attention modules, while in the case of \wmtsmt{}, the position-wise feed-forward layer in $dec_{pe}$ needs to process information from three attention modules, which may increase the learning difficulty of the feed-forward layer.
(3) Since $pe$ is a post-edited version of $mt$, sharing the same language, $mt$ and $pe$ are quite similar compared to $src$. Therefore, attending over a fine-tuned representation from $mt$ along with $src$, which is what we have done in this work, might be a reason for the better results than those achieved by attending over $src$ directly. 

Evaluating the influence of the depth of our encoders and decoder show that while the decoder depth appears to have limited importance, reducing the encoder depth indeed hurts performance which is in line with \newcite{domhan-2018-much}.
    
    

\section{Conclusions}
In this paper, we presented a multi-encoder transformer-based APE model that repurposes the standard transformer blocks in a simple and effective way for the APE task: first, our \textit{transference} architecture uses a transformer encoder block for $src$, followed by a decoder block without masking on $mt$ that effectively acts as a second encoder combining $src \rightarrow mt$, and feeds this representation into a final decoder block generating $pe$. 
The proposed model outperforms the best-performing system of WMT 2018 on the test2016, test2017, dev2018, and test2018 data and provides a new state-of-the-art in APE. 

Taking a departure from traditional transformer-based encoders, which perform self-attention only, our second encoder also performs cross-attention to produce representations for the decoder based on both $src$ and $mt$.
We also show that the encoder plays a more pivotal role than the decoder in transformer-based APE, which could also be the case for transformer-based generation tasks in general.
Our architecture is generic and can be used for any multi-source task, e.g., multi-source translation or summarization, etc.



\bibliography{emnlp-ijcnlp-2019}
\bibliographystyle{acl_natbib}

\end{document}